# Hyperspectral Image Super-Resolution via Dual-domain Network Based on Hybrid Convolution

Tingting Liu , Yuan Liu , Chuncheng Zhang , Liyin Yuan , Xiubao Sui, Qian Chen

*Abstract*—Hyperspectral images (HSIs) with high spatial resolution are hard to obtain directly because of sensor limitations. Deep learning can provide an end-to-end reconstruction solution. Nevertheless, existing methods have two main drawbacks. First, networks with self-attention mechanisms often require a trade-off between internal resolution, model performance and complexity, resulting in the loss of fine-grained, high-resolution features. Second, there are visual discrepancies between the reconstructed hyperspectral image (HSI) and the ground truth because they focus on spatial-spectral domain learning. In this paper, a novel super-resolution algorithm for HSIs, called SRDNet, which uses a dual-domain network with hybrid convolution and progressive upsampling to exploit both spatial-spectral and frequency information of the hyperspectral data, is proposed. In this approach, we design a self-attentive pyramid structure (HSL) to capture interspectral self-similarity in the spatial domain, thereby increasing the receptive range of attention and improving the feature representation of the network. Additionally, we introduce a hyperspectral frequency loss (HFL) with dynamic weighting to optimize the model in the frequency domain and improve the perceptual quality of the HSI, avoiding the over-smoothing caused by spatial loss only. Experimental results on three benchmark datasets show that SRDNet can effectively improve the texture information of the HSI and outperform state-of-the-art methods for HSI reconstruction.

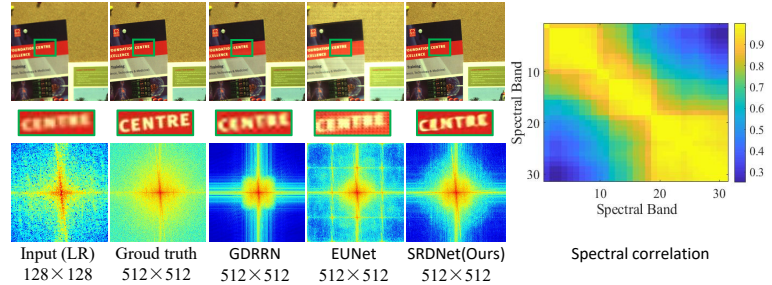

**Fig. 1.** On the left is a comparison of the effects of several reconstruction algorithms, and on the far right is the spectral correlation (Harvard dataset × 4). Correlation values closer to 0 indicate that the two bands are more independent of each other, while correlation values closer to 1 indicate that the two bands are more linearly correlated.

## I. INTRODUCTION

Hyperspectral imaging technology can effectively distinguish different targets or capture the complex spectral details of a scene by analyzing and comparing the spectral features of the images. As a result, hyperspectral imaging (HSI) has many applications in various fields [1-3]. However, a common challenge is the trade-off between spatial and spectral resolution in hyperspectral imagers. To achieve higher spectral resolution, the spatial resolution is often compromised, resulting in low-quality HSI [4, 5]. This low spatial resolution leads to the mixing of end member spectra, which affects the detection performance of the HSI.

Image super-resolution (SR) is the process of recovering high-resolution (HR) images from low-resolution (LR) images, overcoming the resolution constraints of imaging systems [6-8]. Many novel CNN networks have been developed to improve SR performance, using the powerful representation capabilities of CNN. Unlike RGB images, HSI needs to take into account the spectral information in resolution reconstruction (The correlation is illustrated in Fig. 1). Although several CNN-based single image super-resolution (SISR) methods have been developed for RGB images, these algorithms do not effectively utilize the spectral information inherent in HSIs, leading to poor performance [9-11]. HSIs possess a spectral dimension, which can be effectively handled by SISR methods employing 3D convolutions [12-14]. Unfortunately, the use of standard 3D convolution kernels in these methods results in a large number of network parameters.

To address this limitation, subsequent studies have utilized separable convolution kernels, such as $1 \times k \times k$ or $k \times 1 \times 1$, instead of $k \times k \times k$ [13], [15-18]. This approach significantly reduces the number of model parameters and enables the design of deeper networks. In addition, some researchers have employed 2D or 3D separable convolutional kernels to extract spatial-spectral information [9], [19], [12], [20]. Nonetheless, these methods tend to overly emphasize low-frequency pixels in the spatial domain, leading to the blurring of synthesized images.

To address the aforementioned challenges, it is essential to develop a model that considers both spatial-spectral characteristics and frequency characteristics. In light of this, we propose a novel approach called SRDNet for HSI super-resolution. This manuscript makes the following key contributions:

- Considering the spectral dimensionality of the image, it is challenging to capture the complete spatial mapping relationship between low and high resolutions through simple upsampling. To address this, a hybrid convolution of 2D and 3D units with progressive upsampling is designed.
- In the 2D unit, the IGM module is utilized to attain refined and diversified feature expressions, addressing demanding detail capture problems. The IGM module comprises symmetric group convolution block and complementary convolution block, which enhance the internal and external relationships among individual channels in a parallel manner, facilitating the extraction of various types of low-frequency structural information.
- To better use the spectral prior to enhance the learning of spatially global information and spectral coherence, a dual-domain network for SR is designed. Specifically, in the spatial domain, the hyperspectral self-attention learning mechanism (HSL) with pyramidal structure can not only increase the attention field of view, but also model the spectral features with the global spatial spectral context, allowing feature interactions to have different contributions in the reconstruction. In the frequency domain, the hyperspectral frequency loss (HFL) is applied to optimize Fourier transformed model and improve image quality through the dynamic weighting mechanism.

The article is organized as follows: Section 2 provides an overview of related work on HSI SR. In Section 3, we describe the proposed SRDNet method, including the network structure and the dual-domain mechanism. Experimental parameter settings and analysis are presented in Section 4. Lastly, Section 5 concludes the paper.

## II. RELATED WORK

Some of the works most relevant to our approach are briefly reviewed, including CNN-based methods, attentional mechanisms, and the use of frequency domain analysis in SISR.

### A. Single HSI Super-resolution

Previous studies mainly formulated HSI SR as a constrained optimization problem, where some priors were used to limit the space [21]. For instance, to explore the non-local similarity in the spatial domain, Wang et al. [22] modeled three properties of the HSI: the non-local similarity in the spatial domain, the global correlation in the spectral domain, and the smooth structure of the spatial-spectral domain [23]. Huang et al. [24] used sparse and low-rank properties to reconstruct HSI with spatial super-resolution. Recently, CNNs have become more popular than traditional optimization-based solutions for SR, due to their strong feature extraction and representation abilities. Xie et al. [25] and Yuan et al. [26] employed DCNN networks with non-negative matrix decomposition to preserve the spectral properties of the intermediate results for HSI resolution. However, DCNN networks have difficulty in fully exploiting the spatial and spectral properties of HSI, due to the large number of spectral dimensions and the lack of enough training samples. Li et al. [9] proposed a grouping strategy, called the GDDRN method with a recursive module, to better capture the correlation between spectral bands. The method combines a spectral angle mapper (SAM) with mean squared error (MSE). However, its loss function affects the enhancement of spatial resolution. Mei et al. [12] developed a neural network, named 3D-FCNN, which used 3D convolution to jointly explore spatial texture features and spectral correlations. Although it reduces the distortion of the spectrum, it is computationally expensive at a high upsampling factor. Hu et al. [27] integrated a collaborative non-negative matrix factorization (CNMF) strategy with the outputs of a deep feature extraction network for learning. The method works well, but it considers spatial and spectral features separately and relies too much on manual intervention. Zheng et al. [11] designed a separable spectral convolution to obtain information on each spectral band. However, this would generate many redundant features with hundreds of spectral bands, making it hard to identify the most representative ones. Despite the good performance of EUNet [28] proposed by Liu et al. with a small number of parameters, it is not the optimal choice. This method fails to fully utilize spectral information and instead focuses excessively on extracting spatial features, resulting in poor SAM scores. Li et al. [6] proposed a network with an attention mechanism, which is called 3D generative adversarial network to mitigate the issue of inter- spectral distortion. However, regular 3D convolution often has high storage complexity and computation cost. To better extract spatial and spectral features, Li et al. [14] designed a hybrid network by combining the 2D convolution and the 3D separable convolution. But it is computationally expensive, and the network structure causes information redundancy.

These HSI SR methods have achieved remarkable results, but they have some limitations. On one hand, the existing methods focus on learning the spatial-spectral information of HSI from the spatial domain. We address the challenge of synthesizing high frequencies by applying a separate module to the SR results. This module adaptively recovers high and hard frequencies, resulting in a higher resolution of the internal features. On the other hand, some algorithms do not selectively concentrate on important features and lack a global context to model spectral dependencies. Some features in certain locations and channels are more useful for SR reconstruction. To utilize the features more effectively, we devise a novel spatial-spectral self-attention mechanism in our method, which can acquire more detailed information and suppress other irrelevant information.

*B. Attention Mechanism*

The attention mechanism is a signal processing mechanism that scans the image and allocates more attention to the focus of attention. It attends to the details of the target and suppresses other irrelevant information [29]. The attention mechanisms can be used in SR tasks to concentrate on prominent information, reduce image noise and improve the quality of reconstructed images. Some SR methods use attention mechanisms from other vision tasks [30, 31] to focus more on spectral characteristics. Dai et al. introduced second-order channels to capture long-distance spatial differences between SR and other tasks. Moreover, the attention mechanisms have been applied to HSI SR because of their powerful representation abilities [6, 32]. But most existing methods ignore the internal resolution of attention to speed up computations, which leads to degraded algorithm performance. There are some pixel-level attention mech-anisms that are designed for advanced tasks [33-35] that further improve the representation abilities of the model. It is important to explore a pixel-level attention mechanism that can increase the effectiveness of HSI reconstruction.

*C. Image Frequency Domain Analysis*

Spectrum analysis can decompose a complex signal into simpler signals. F-Principle [36] showed that deep learning methods tend to focus on low frequency to reconstruct targets, which causes differences in the frequency domain. In recent years, many CNN-based methods have been proposed to analyze the frequency domain. A coordinate-based MLP with Fourier transform [37] was used to recover high frequencies missing in single image reconstruction. Recent works have shown that frequency analysis can be integrated with SR [38, 39]. For example, some works have tried to reconstruct better visual images by minimizing the frequency domain difference between input and output during training [40]. In HSI SR, where the model is more likely to concentrate on low-frequency pixels in the spatial domain, the composite image becomes more blurred. Therefore, exploring an adaptive constraint based on intrinsic frequency is essential for reconstructing fine images.

### III. THE PROPOSED METHOD

*A. Overall Architecture*

In this subsection, we present the detailed overall architecture of SRDNet, and show its algorithm model diagram in Fig. 2. Our method consists of four main components: the shallow feature extraction, the deep spatial-spectral feature extraction, the upsampling, and the reconstruction part. For HSI SR, we denote the LR of input and the reconstructed SR image as $I_{LR} \in \mathbb{R}^{B \times H \times W}$ and $I_{SR} \in \mathbb{R}^{B \times rH \times rW}$, respectively. The $W$ and $H$ are the width and height of the HSI, and $B$ is the number of spectral bands in the HSI. The $r$ is the SR scale factor that is generated by the LR. Our aim is to reconstruct the SR($I_{SR}$) end-to-end by the network (SRDNet) from the LR($I_{LR}$), as shown in Eq. (1).

$$I_{SR} = H_{Net}(I_{LR}) \quad (1)$$

where $H_{Net}(.)$ denotes the function of the proposed super-resolution network. The flow of the overall network architecture is as follows.

Firstly, the convolutional layer and the first residual block are defined as $F_{conv}(.)$ and $F_{Res}^{(1)}(.)$, respectively, through which the shallow features are extracted. The corresponding feature $x_0$ is defined as:

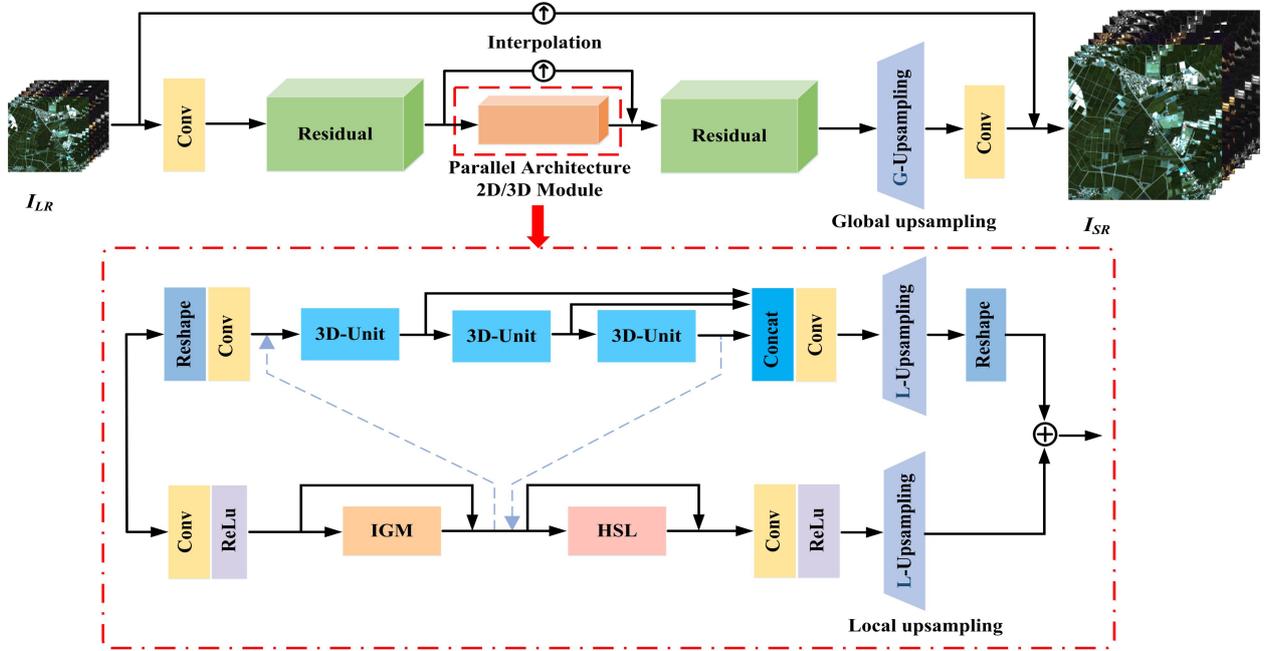

**Fig. 2.** Overall architecture of the SRDNet

$$x_0 = F_{\text{Res}}^{(1)}(F_{conv}(I_{LR})) \quad (2)$$

Secondly, the parallel structure 2D/3D module (PAM) and the second residual block $F_{\text{Res}}^{(2)}(.)$ are adopted to extract the deep features. The corresponding feature $x_t$ is defined as:

$$x_t = F_{\text{Res}}^{(2)}(F_{PAM}(x_0) + F_{L\_up}(x_0 \uparrow, r)) \quad (3)$$

Finally, there is the upsampling and reconstruction part, in order to upgrade the acquired features to the target size, where the upsampling module is introduced to generate a spatial spectral feature map of the target. To alleviate the affordability of the final SR reconstruction, the strategy of progressive upsampling is applied in this paper, which is divided into local upsampling and global upsampling. The $F_{L\_up}(x_0\uparrow)$ denotes the local bicubic operation. The corresponding feature $x_{up}$ is defined as:

$$x_{rec} = F_{conv}(F_{G\_up}(x_t, r)) \quad (4)$$

Where $x_{rec}$ is represented by convolutional layer reconstruction, and $F_{G\_up}(.)$ denotes the global upsampling operation, and the transposed 2D convolutional layer is applied to upsample the feature map by the scale factor $r$.

$$I_{SR} = x_{rec} + F_{up}(I_{LR}\uparrow, r) \quad (5)$$

Where $F_{up}(I_{LR}\uparrow)$ represents the input $I_{LR}$ for up-sampling operation by bicubic interpolation. Equation (5) is equivalent to Eq. (1).

The process of the overall network architecture has been described above. Next, we will introduce the network sub-modules in both the spatial-spectral domain and the frequency domain, respectively. The parallel architecture 2D/3D module will be elaborated from the spatial domain. In addition, the hyperspectral frequency loss (HFL) will be described in detail from the frequency domain.

*B. Parallel architecture 2D/3D module (PAM)*

After shallow feature extraction, the information flow is divided into two branches, which the first is a 3D convolutional branching sub-network and the second is a 2D convolutional branching sub-network.

**1) 3D Unit Branch Network**

As mentioned in the second subsection, HSIs have extra spectral dimension compared to RGB images, which allows the use of 3D convolution to gain information beyond the spatial dimension. This paper makes use of separable 3D convolution (the convolution kernel of filter k×k×k is replaced with k × 1 × 1 and 1 × k × k) instead of the conventional 3D convolution. To be able to use 3D convolution, since the size of the input HSI is B × W × H, the $I_{LR}$ is reshaped into four dimensions (C×B×W×H), and C is the number of channels.

$$y_0^{3D} = f_{1\times1\times1}^{3D}(f_{Reshape}(x_0 + y_{IGM}^{2D})) \quad (6)$$

where the $y_{IGM}^{2D}$ repents the output of the IGM unit, $f_{Reshape}(.)$ denotes the extended dimension of the feature map that extracts from the first residual group block. As for the 3D unit, as shown in Fig. 3. The kernel is the same for each 3D unit, and in the case of the first 3D unit, the output can be represented as follows.

$$y_{1D}^{3D} = \sigma(f_{3\times1\times1}^{3D}(\sigma(f_{1\times3\times3}^{3D}(y_0^{3D})))) \quad (7)$$

The output of the previous 3D unit is employed as input to the next 3D unit. Assuming that there are *N* 3D units in the model, the output is shown as:

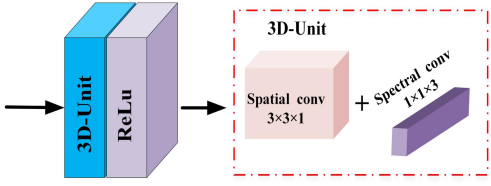

**Fig. 3.** Architecture of the 3D-unit

$$y_{ND}^{3D} = f_{ND}^{3D}(f_{(N-1)D}^{3D}(\cdots f_{2D}^{3D}(f_{1D}^{3D}(y_0^{3D}))\cdots)) \quad (8)$$

Where $f_{ND}^{3D}(.)$ is operation of the $N_{th}$ 3D unit (in order to make the network simple and effective, here $N=3$).

Once the hierarchical features have been obtained from $N$ 3D units, their outputs are combined or cascaded together to enable the network to capture more valuable information. Subsequently, local upsampling is carried out, and the corresponding feature operations are as follows.

$$y_{L\_up}^{3D} = f_{L\_up}^{3D}(\sigma(f_{1\times1\times1}^{3D}(f_{Concat}^{3D}[y_{1D}^{3D}, y_{2D}^{3D}, y_{1D}^{3D}]))) \quad (9)$$

Among them, the $f_{L\_up}^{3D}(.)$ is represented a local upsampling operation. Finally, the obtained feature $y_{L\_up}^{3D}$ is reshaped from four dimensions (C × B × W × H) to the image size of three dimensions (B×W×H). The output of the 3D Unit branch is obtained.

$$y^{3D} = f_{Reshape}(y_{L\_up}^{3D}) \quad (10)$$

Where $y^{3D}$ is the dimension of the feature being squeezed.

**2) 2D Unit Branch Network**

In contrast to the 3D unit operation, the 2D unit branching network is used to extract deeper features through the IGM unit. Its structure is shown in Fig. 4.

**a) Isomeric Group Module (IGM)**

Previous HSI SR work directly merges the layered features of all channels to enhance the image resolution performance, which may increase the redundant features and the convergence time of the algorithm. To solve this issue, we use the IGM module by interacting various channels to obtain deeper low-frequency information, which strengthens the relationship between different channels and improves the performance of the SR algorithm. The internal and external relationships of the various channels are improved by the parallel nature of acquiring different types of more representative structural information.

Symmetric Group Convolution Block: This block consists of two subnetworks with three layers each, which extract the main channel information from the image. The features obtained by the subnetworks are combined by a cascading operation to enhance their internal coherence. Each layer of the subnetworks follows the Conv+ReLU structure, with 32 input and output channels and 3×3 convolution kernels. The output of the symmetric group block is obtained by cascading the outputs of the two subnetworks, resulting in 64 output channels. The two subnetworks have identical structures. For example, the output of the first subnetwork is shown as follows:

$$x_0^{2D} = \sigma(f_{conv}^{3\times3}(x_0)) \quad (11)$$

$$y_{1b}^{2D} = \sigma(f_{conv}^{3\times3}(x_0)) \quad (12)$$

$$y_{SGB-1}^{2D} = f_{3b}^{2D}(f_{2b}^{2D}(y_{1b}^{2D})) \quad (13)$$

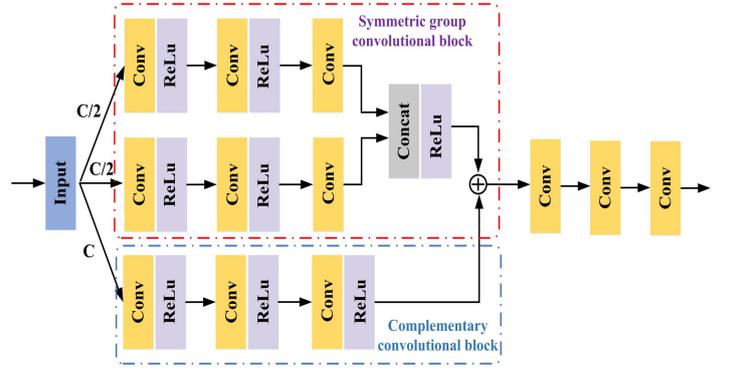

**Fig. 4.** IGM structure diagram

Where the $f_b^{2D}(.)$ denotes the operation of convolution and ReLU function per layer, for a total of 3 layers. The second branch subnet has the same operation as the first, and the output feature representation is as follows:

$$y_{SGB-2}^{2D} = f_{3b}^{2D}(f_{2b}^{2D}(y_{1b}^{2D})) \quad (14)$$

The output features of the symmetric group convolutional block are obtained by cascading the output features of the two subnetworks, which is the output feature representation of this network.

$$y_{SGB}^{2D} = f_{Concat}^{2D}[y_{SGB-1}^{2D}, y_{SGB-2}^{2D}] \quad (15)$$

Complementary Convolution Block: This block consists of a single three-layer subnetwork that captures the total information of all channels in the image. The complementary convolution block improves the external correlation and robustness of the algorithm, which is a valuable complement to the symmetric group convolution block. Each layer of the complementary convolution block has the same structure as the symmetric group convolution block. The difference is that each convolution layer has 64 input, which allows for more fine-grained feature extraction. The output of the complementary convolution block is shown as:

$$y_{CB}^{2D} = f_{3b}^{2D}(f_{2b}^{2D}(y_{1b}^{2D})) \quad (16)$$

The output of symmetric group convolution blocks and complementary convolution blocks are spatially superimposed and integrated by a convolution fusion operation, which produces the output of the IGM module. The output characte-

ristic of this module is expressed as:

$$y_{IGM}^{2D} = f_{conv}(y_{CB}^{2D} + y_{SGB}^{2D}) + x_0^{2D} \quad (17)$$

b) HSL module

To begin with, the indiscriminate treatment of all types of features in the convolution kernel limits the ability of the network to reconstruct. It is difficult to distinguish useful high-frequency information from rich low-frequency features [41]. Moreover, the perceptual field of each layer of the convolutional kernel is only a local expression of the fact that the output is only partially related to the input, and it lacks sufficient contextual information. Furthermore, the spectral features are of great importance for the HSI, so an adequate exploration of spectral correlations cannot be neglected. Based on the above three considerations, we incorporate attention mechanisms from the spatial and spectral dimensions, respectively. This allows the exploration of the long-range dependence of pixel positions and the correlation of spectral bands. The specific structure is illustrated in Fig. 5.

Spectral attention: The convolutional layer is utilized to linearly map the input feature to obtain the **query** vector and **key** vector. And the full space **query** vector and half channel **key** vector are obtained by reshape operation. Then, the **query** vector is remapped to the **key** vector through the matrix multiplication. And the auto-correlation of the spectral dimension is calculated to obtain the attention coefficient, which is recorded as the **value** vector. Its channel is maintained as C/2, which avoids excessive computational costs.

$$y_{Spectal-q}^{2D} = f_{Softmax}(f_{Reshape}(f_{conv}(y_{IGM}^{2D} + y_{3D}^{3D}))) \in \mathbf{R}^{1,HW} \quad (18)$$

$$y_{Spectal-k}^{2D} = f_{Reshape}(f_{conv}(y_{IGM}^{2D} + y_{3D}^{3D})) \in \mathbf{R}^{C/2,HW} \quad (19)$$

$$y_{Spectral-v}^{2D} = y_{Spectral-q} \otimes y_{Spectral-k}^{2D} \in \mathbf{R}^{C/2,1,1} \quad (20)$$

where the $\otimes$ denotes the matrix multiplication operation and the $f_{Softmax}(.)$ represents the softmax function. As shown in the Fig. 4, after the value vector is activated by convolution and sigmoid, weight coefficient of each channel can be obtained.

$$y_{Spectral-map}^{2D} = f_{Sigmoid}(f_{conv}(y_{Spectral-v}^{2D})) \in \mathbf{R}^{C,1,1} \quad (21)$$

where $f_{Sigmoid}(.)$ denotes the sigmoid activation function and $f_{Spectral-map}^{2D}(.)$ denotes the channel weight coefficient.

Finally, the $y_{IGM}^{2D}$ is multiplied by the weight factor of each channel to obtain the calibrated input signal, so that the HSI focuses on the channel with the greater weight.

$$y_{Spectral}^{2D} = y_{IGM}^{2D} \odot y_{Spectral-map}^{2D} \in \mathbf{R}^{C,H,W} \quad (22)$$

Spatial attention: To enhance the fused features with spatial awareness, we compute the spatial attention mask based on spectral attention. A pyramid structure was designed to progressively expand the attentional receptive field to capture the spatial context at different scales. Then, we perform element-wise multiplication and addition operations to combine the features with the mask to fuse the features in a more adaptive and effective way.

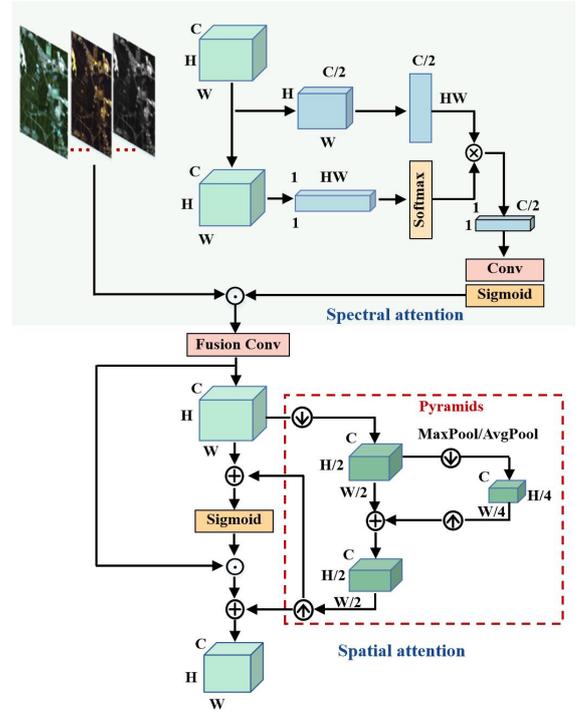

**Fig. 5.** Hyperspectral attention mechanism

$$y_{Fusion}^{2D} = f_{conv}(y_{Spectral}^{2D}) \in \mathbf{R}^{C,H,W} \quad (23)$$

$$y_{Down}^{2D} = f_{conv}[y_{max}^{2D}, y_{avg}^{2D}] \in \mathbf{R}^{C, H/2, W/2} \quad (24)$$

$$y_{Pyramid}^{2D} = f_{Up}(f_{Down}(y_{Fusion}^{2D})) + y_{Down} \in \mathbf{R}^{C,H/2, W/2} \quad (25)$$

Where the $y_{Fusion}^{2D}$ denotes the fusion of spectrally corrected features, the $y_{max}(.)$ and $y_{avg}(.)$ denote the maximum pooling and average pooling operations of downsampling, respectively. The $f_{Down}(.)$ denotes the acquired post-sampling features and the $f_{Up}(.)$ denotes the upsampling operation.

As shown in Fig. 5, the obtained features have undergone two downsampling and upsampling respectively. Then, according to the sigmoid function, the weight coefficient of the spatial pixel can be obtained.

$$y_{Spatial-map}^{2D} = f_{Sigmoid}(f_{Up}(y_{Pyramid}^{2D}) + y_{Fusion}) \in \mathbf{R}^{C,H,W} \quad (26)$$

$$y_{Spatial}^{2D} = y_{Fusion} \odot y_{Spatial-map}^{2D} \in \mathbf{R}^{C,H,W} \quad (27)$$

Where the $f_{Spatial-map}^{2D}(.)$ denotes the obtained spatial pixel weighting factor. After the calibrated input signal is obtained, so that the HSI focuses on the pixel area of greater weight.

The corrected feature signals in the spectral and spatial domains are fused so that the output feature maps are highly correlated in both spatial pixels and spectral dimensions.

$$y^{2D}_{HSL} = (y^{2D}_{Spectral} + y^{2D}_{Spatial}) + y^{2D}_{IGM} \quad (28)$$

where the $f^{2D}_{HSL}$ denotes the output characteristics of the HSL module. Similar to the 3D branch network, local upsampling is used to alleviate the burden of the final SR reconstruction. This is shown in the following operations:

$$y^{2D} = f^{2D}_{L\_up}(\sigma(f_{conv}(y^{2D}_{HSL}))) \quad (29)$$

In the PAM network, the outputs of the 3D unit branch network and the 2D unit branch network are merged from spatial pixels. The definition is shown below.

$$x_t = y^{2D} + y^{3D} \quad (30)$$

where the $x_t$ denotes the output of the entire PAM block resulting from the result of the operation of the $F_{PAM}(\cdot)$ function in Eq. (3)

c) Hyperspectral frequency loss (HFL)

Discrete Fourier transform (DFT) image analysis: In order to process and analyze the image in the frequency domain, we apply a two-dimensional discrete Fourier transform (DFT) to convert the HSI from a spatial domain to a frequency domain representation, as shown in Figure 6(b).

$$F_{(u,v)} = \sum_{x=0}^{H-1}\sum_{y=0}^{W-1} f(x,y) \cdot e^{-i2\pi(\frac{ux}{H}+\frac{vy}{W})} \quad (31)$$

where the image dimensions are H×W; the (x, y) denotes the coordinate of the image pixels in the spatial domain; and the $f(x,y)$ is a pixel value. The $(u,v)$ denotes the spatial coordinate of the spectrum; the $F(u,v)$ is the complex frequency value, which is the sum of each image pixel in the spatial domain.

HSI frequency distance analysis: We can use the distance metric to measure the difference between the reconstructed SR and the ground truth in the frequency domain, as shown in Fig. 6(a). In the frequency domain, we operate on frequencies in the same space, which are represented as different 2D sinusoidal components in the image. We need to take into account both amplitude and phase when using frequency distances, as they capture different features of the image. We map each frequency value into a two-dimensional space (i.e., a plane) and convert it into a Euclidean vector. The frequency distance is then computed as the distance between the $\vec{r}_{GT}$ and the $\vec{r}_{SR}$, which involves both the angle and magnitude of the vector (the purple line in Fig. 6(a)). We use the squared Euclidean distance for a single frequency (the $k_{th}$ example).

$$d^k_{(\vec{r}_{GT},\vec{r}_{SR})} = \|\vec{r}^k_{GT} - \vec{r}^k_{SR}\|^2_2 = |F^k_{GT}(u,v) - F^k_{SR}(u,v)|^2 \quad (32)$$

Where k= {0, 1, 2, ..., C-1}, and the frequency distance between the ground truth and the reconstructed image can be written as the mean value.

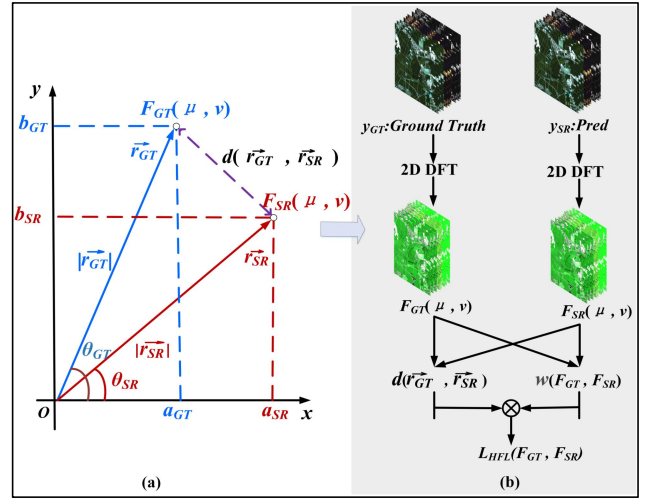

**Fig. 6.** Hyperspectral frequency loss

$$d^k_{(u,v)} = \frac{1}{HW}\sum_{u=0}^{H-1}\sum_{v=0}^{W-1} |F^k_{GT}(u,v) - F^k_{SR}(u,v)|^2 \quad (33)$$

Due to the inherent bias, the network will still be biased towards easy frequencies, so to achieve the training focus on hard frequencies, a spectral weight matrix $w(u,v)$ is introduced to weight the easy frequencies. The specific definition is as follows:

$$w^k_{(u,v)} = |F^k_{GT}(u,v) - F^k_{SR}(u,v)|^\alpha \quad (34)$$

where α is the scaling factor for flexibility (in the experiment α=1). Further normalize w(u,v) to the range [0,1], where the 1 corresponds to the frequency that is currently lost the most and the easiest. Each frequency has a different weights.

Taking the $w(u,v)$ matrix value and the frequency distance matrix by Hadamard (shown in Fig. 6(b)), we get the output of the HSI frequency loss (HFL) as follows.

$$d(F^k_{GT},F^k_{SR}) = \frac{1}{HW}\sum_{u=0}^{H-1}\sum_{v=0}^{W-1} w^k_{(u,v)} |F^k_{GT}(u,v) - F^k_{SR}(u,v)|^2 \quad (35)$$

$$L_{HFL}(F_{GT},F_{SR}) = \sum_{k=0}^{C-1} d(F^k_{GT},F^k_{SR}) \quad (36)$$

d) Total training losses

The $L_1$ loss is a common choice for SR works, as it can ensure good convergence in training. In this article, we use the $L_1$ loss to compute the pixel loss in the spatial domain. At the same time, the $L_{HFL}$ (shown in Eq. 36) is used to calculate the loss in the spectral frequency domain.

$$L_1(\Theta) = \frac{1}{N}\sum_{n=1}^{N} \|I^n_{GT} - H_{Net}(I^n_{LR})\|_1 \quad (37)$$

Where the $I^n_{GT}$ and the $H_{Net}(I^n_{LR})$ denote the $n_{th}$ ground truth and the reconstructed hyperspectral image, the $N$ is indicated the number of HSI in a training batch, and the $\Theta$ is used as the parameter set of our network. The total loss $L_{total}$ combined with the dual-domain network is expressed as:

$$L_{total} = L_1 + \beta L_{HFL} \quad (38)$$

Where the $\beta$ is utilized to balance the contribution of various losses, and the $\beta$ is set to 0.1.

## IV. EXPERIMENTS

### A. Experiment Settings

1) Datasets

a) CAVE database: The CAVE database[1] was collected with a cooled CCD camera [42] in wavelength range of 400nm-700nm (31 bands) in 10 nm steps. The database contains 32 object scenes with the HSI sizes of 512×512×31.

b) Harvard database: The Harvard dataset[2] was acquired using Nuance FX, CRI Inc.cameras [43] and was located in a scene of daylight or outdoor scene. The dataset contains 77 HSIs, each with a size of 1040×1392×31.

c) Chikusei database: The Chikusei dataset[3] was captured by the Headwall Hyperspec-VNIR-C sensor in Chikusei, Japan. It encompasses a wavelength range of 343 nm to 1018 nm and comprises 128 spectral bands. The HSIs have a spatial resolution of 2.5 meters and a size of 2517×2335.

2) Implementation Details

Each dataset was captured using a different camera, requiring separate training and testing. In experiments, 80% of the datasets were randomly allocated for training, 10% for validation, and 10% for testing. During training, we randomly selected 24 patches for augmenting the training data. These patches underwent operations such as horizontal flipping (90°, 180°, 270°) and scaling (1, 0.5, 0.75). Subsequently, the patches were downsampled using interpolation with a scale factor of $r$, resulting in low spatial resolution images sized 32×32×C. The initial learning rate is set to 1.0e-4, and it has a total of 200 epochs. Our experiments were conducted on the Ubuntu 18.04 operating system, utilizing the PyTorch 1.7.1 deep learning framework. Hardware acceleration was performed using an RTX 3090 GPU. To optimize test efficiency, we selected a 512×512 area in the upper left corner as the test image for evaluation.

3) Evaluation Metrics

To assess the performance of our network, we used four popular quantitative measures of image quality (PQI): peak signal-to-noise ratio (PSNR) and structural similarity (SSIM)

---
[1] https://www.cs.columbia.edu/CAVE/databases/
[2] http://vision.seas.harvard.edu/hyperspec/
[3] https://naotoyokoya.Com/Download.html

[44], mutual correlation (CC) [45], and spectral angle mapper (SAM) [46]. PSNR and SSIM are common metrics for image restoration quality, and their ideal values are $+\infty$ and 1, respectively. CC and SAM are frequently used in HSI fusion works, and their best values are 1 and 0, respectively.

### B. Comparisons With the State-of-the-Art Methods

We conducted a thorough comparison of the SRDNet with six classical methods: Bicubic, GDRRN [9], 3D-FCNN [12], EDSR [47], EUNet [28], and MCNet [14]. Three datasets, namely CAVE, Harvard, and Chikusei, were utilized to validate the benefits of SRDNet across various sampling factors and bands' numbers.

1) Results on CAVE Dataset

TABLE I
COMPARISON OF ALGORITHMS IN TERMS OF NUMBER OF CAVE PARAMETERS

| Methods | ×2 | ×3 | ×4 |
|---|---|---|---|
| Bicubic | - | - | - |
| GDRRN[9] | 0.22M | 0.22M | 0.22M |
| 3D-FCNN[12] | 0.039M | 0.039M | 0.039M |
| EDSR[47] | 1.40M | 1.59M | 1.55M |
| MCNet[14] | 1.93M | 2.04M | 2.17M |
| EUNet[28] | 0.56M | 0.63M | 0.62M |
| SRDNet(Ours) | 1.57M | 1.72M | 1.60M |

TABLE II
QUANTITATIVE COMPARISON OF DIFFERENT METHODS FOR THE 4 PQIS PAIRS OF 5 TEST IMAGES ON THE CAVE DATASET. RED AND BLUE SHOW THE BEST AND SECOND BEST PERFORMANCE RESPECTIVELY.

| Methods | Scale | CC↑ | SAM↓ | PSNR↑ | SSIM↑ |
|---|---|---|---|---|---|
| Bicubic | ×2 | 0.990 | 2.605 | 42.662 | 0.9885 |
| GDRRN[9] | | 0.991 | 2.609 | 43.423 | 0.9898 |
| 3D-FCNN[12] | | 0.991 | 2.605 | 43.377 | 0.9864 |
| EDSR[47] | | 0.991 | 2.905 | 44.362 | 0.9903 |
| MCNet[14] | | 0.992 | 2.417 | 45.082 | 0.9915 |
| EUNet[28] | | 0.992 | 2.541 | 45.261 | 0.9921 |
| SRDNet(Ours) | | 0.993 | 2.214 | 46.424 | 0.9929 |
| Bicubic | ×3 | 0.983 | 3.255 | 38.988 | 0.9763 |
| GDRRN[9] | | 0.984 | 2.895 | 39.337 | 0.9788 |
| 3D-FCNN[12] | | 0.985 | 2.907 | 41.345 | 0.9792 |
| EDSR[47] | | 0.988 | 3.163 | 42.432 | 0.9859 |
| MCNet[14] | | 0.988 | 2.750 | 42.785 | 0.9863 |
| EUNet[28] | | 0.989 | 4.419 | 42.356 | 0.9862 |
| SRDNet(Ours) | | 0.989 | 2.672 | 42.944 | 0.9869 |
| Bicubic | ×4 | 0.979 | 3.654 | 36.853 | 0.9638 |
| GDRRN[9] | | 0.980 | 3.622 | 37.164 | 0.9649 |
| 3D-FCNN[12] | | 0.981 | 3.421 | 38.471 | 0.9683 |
| EDSR[47] | | 0.985 | 3.258 | 40.135 | 0.9784 |
| MCNet[14] | | 0.986 | 3.033 | 40.271 | 0.9801 |
| EUNet[28] | | 0.991 | 5.324 | 40.308 | 0.9802 |
| SRDNet(Ours) | | 0.986 | 3.025 | 40.612 | 0.9807 |

As shown in Table I, our method has slightly more parameters than EDSR, but it achieves a better result under the same scale factor. Figure 7 shows the visual quality of the reconstructed images for each algorithm. And figures 7 and 8 show the difference in the reconstruction information between the algorithms using error plots and spectral fitting plots. Table II and figure 7 demonstrate that all algorithms perform well in reconstructing high-resolution HSIs. Specifically, the proposed SRDNet method surpasses the spatially-based prior EDSR [47] and the spectral prior 3D-FCNN [12]. GDRRN [9] and EUNet algorithms are designed for (HSI), but their performance on different datasets is unstable. In particular, the SAM value of the EUNet [28] algorithm performs poorly at upsampling factors of 3 and 4. In Table II, SRDNet algorithmexhibits higher PSNR and SSIM values compared to the second-best method, with improvements of 1.163 dB/ 0.0008, 0.159 dB/0.0006, and 0.304 dB/0.0005, respectively, for the upsampling factors of 2/3/4. Notably, the PSNR shows a significant improvement of 2.6% for an upsampling factor of 2. Additionally, as indicated in Table I, the SRDNet method has a smaller parameter count compared to MCNNet.

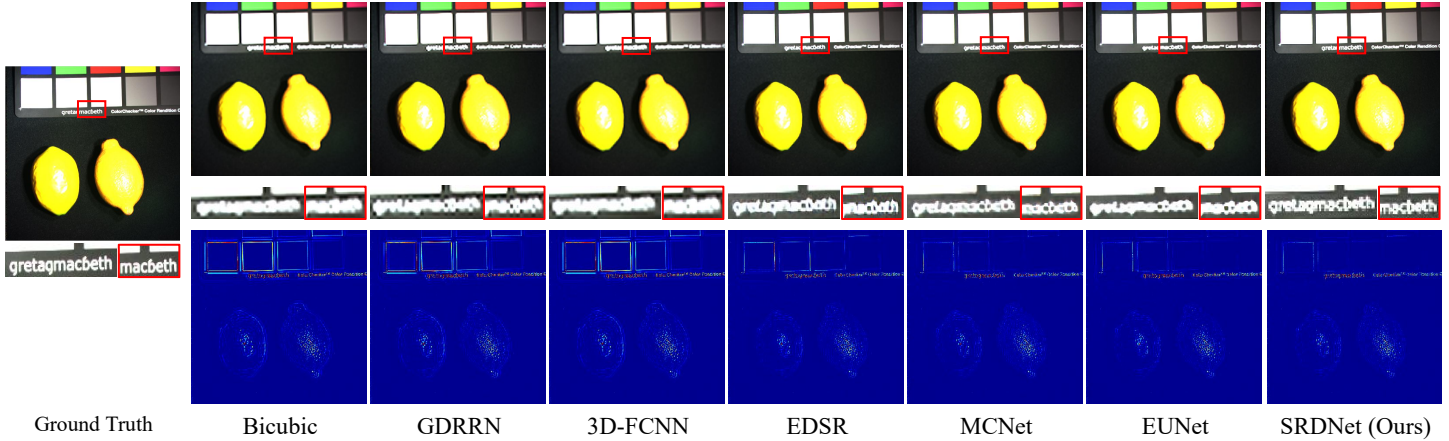

Ground Truth | Bicubic | GDRRN | 3D-FCNN | EDSR | MCNet | EUNet | SRDNet (Ours)

**Fig. 7** The first row depicts the reconstructed *fake_and_real_lemons_ms* image from the CAVE dataset, using spectral bands 23-15-8 as R-G-B, with an upsampling factor of 4. The second row represents the absolute error, indicating that the reconstructed image approaches the target image more closely when it contains less information.

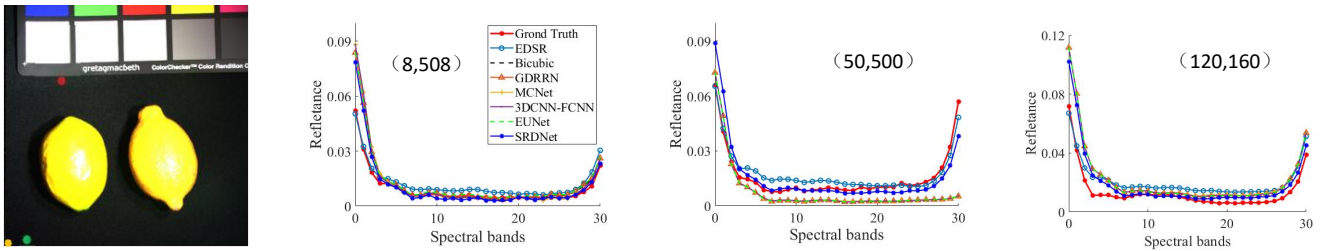

**Fig. 8.** The visual comparison of the *fake_and_real_lemons_ms* image spectral distortion at pixel locations (8,508), (50,500) and (120,160) on the CAVE dataset.

2) Results on Harvard Dataset

Results on Harvard Datas According to Table III, the SRDNet algorithm achieves favorable results across all metrics, similar to the CAVE dataset. The absolute error plot and the spectral fitting plot are utilized to highlight the disparities in edge information between the ground truth and the reconstructed HSI. The Fig. 9 illustrates the spectrogram of each algorithm, where the green background represents high-frequency information and the center region indicates low-frequency information. Bicubic method tends to lose high-frequency details during image upscaling, resulting in blurry and fuzzy enlarged images. In contrast, neural network-based super-resolution algorithms, such as MCNet [14] and SRDNet, preserve high-frequency information better and produce sharper enlarged images. Moreover, Fig. 9 demonstrates that combines both 2D and 3D modules performs better than those with only a 2D or 3D module. This is because 3D convolution effectively utilizes the frequency spectrum for enhanced feature extraction, while 2D convolution improves spatial resolution.

Table III and Fig. 11 present the average performance of different algorithms on the Harvard dataset. SRDNet outperforms the other methods for all three upsampling factors. Particularly, our algorithm, in the case of *imgd6*, effectively preserves image edges during the reconstruction process. For upsampling factors of 2/3/4, our algorithm achieves higher PSNR and SSIM values compared to the second-best method, with improvements of 0.260 dB/ 0.0012, 0.209 dB/0.0009, and 0.061 dB/ 0.0009, respectively.

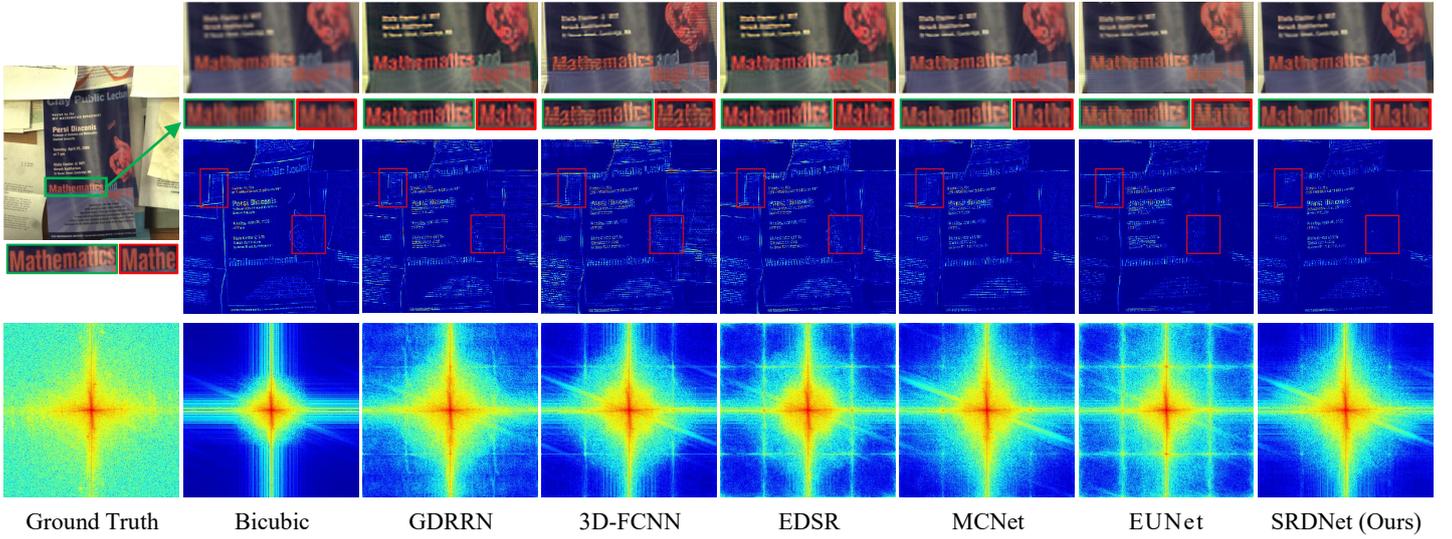

**Fig. 9** The first row illustrates the reconstructed *img3* image from the Harvard dataset using spectral bands 23-15-8 as R-G-B with an upsampling factor of 4. The second row displays the absolute error, demonstrating that the reconstructed image becomes closer to the target image when it contains less information. The third row represents the spectrograms.

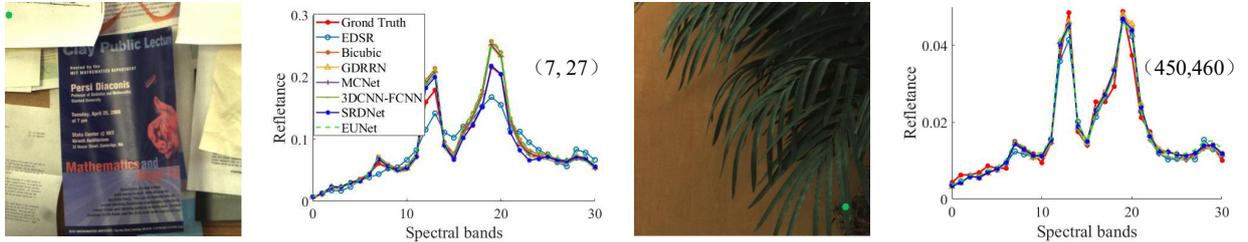

**Fig. 10.** This is a visual comparison of the image spectral distortion on the Harvard dataset.

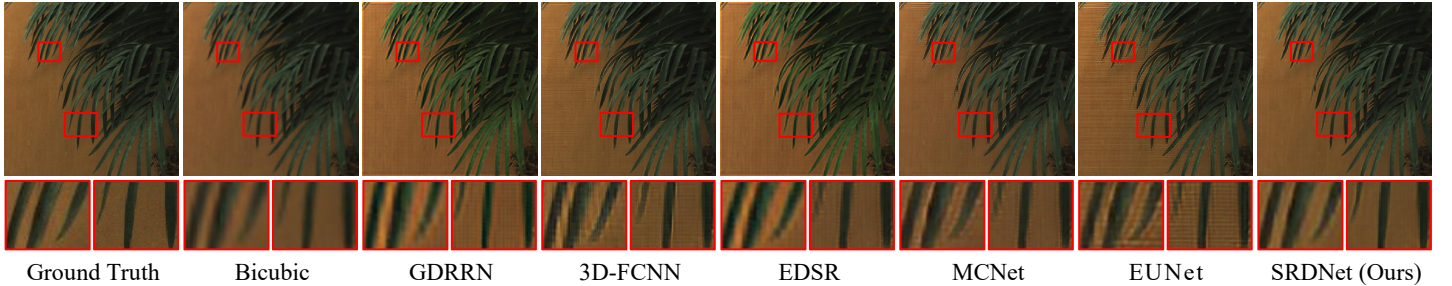

**Fig. 11.** Reconstructed composite images of *imgd6* in Harvard dataset with spectral bands 23-15-8 as R-G-B when the upsampling factor is 4.

TABLE III
QUANTITATIVE EVALUATION OF 25 TEST IMAGES FOR 4 PQIS ON HARVARD DATA. RED AND BLUE SHOW THE BEST AND SECOND BEST PERFORMANCE RESPECTIVELY.

| Methods | Scale | CC↑ | SAM↓ | PSNR↑ | SSIM↑ |
|---|---|---|---|---|---|
| Bicubic | | 0.983 | 3.091 | 42.557 | 0.9728 |
| GDRRN[9] | | 0.998 | 2.284 | 45.300 | 0.9889 |
| 3D-FCNN[12] | | 0.998 | 2.313 | 45.620 | 0.9874 |
| EDSR[47] | ×2 | 0.997 | 2.867 | 44.557 | 0.9808 |
| MCNet[14] | | 0.998 | 2.236 | 46.845 | 0.9893 |
| EUNet[28] | | 0.991 | 2.355 | 46.364 | 0.9840 |
| SRDNet(Ours) | | 0.998 | 2.218 | 47.105 | 0.9905 |
| Bicubic | | 0.979 | 4.262 | 38.714 | 0.9485 |
| GDRRN[9] | | 0.995 | 3.214 | 41.070 | 0.9712 |
| 3D-FCNN[12] | | 0.994 | 3.073 | 40.839 | 0.9690 |
| EDSR[47] | ×3 | 0.994 | 3.914 | 40.864 | 0.9691 |
| MCNet[14] | | 0.997 | 2.549 | 43.742 | 0.9837 |
| EUNet[28] | | 0.986 | 6.295 | 41.519 | 0.9811 |
| SRDNet(Ours) | | 0.997 | 2.443 | 43.951 | 0.9846 |
| Bicubic | | 0.976 | 4.477 | 37.815 | 0.9397 |
| GDRRN[9] | | 0.994 | 3.206 | 38.631 | 0.9669 |
| 3D-FCNN[12] | | 0.994 | 3.169 | 38.692 | 0.9675 |
| EDSR[47] | ×4 | 0.991 | 4.236 | 38.407 | 0.9651 |
| MCNet[14] | | 0.996 | 2.770 | 42.841 | 0.9804 |
| EUNet[28] | | 0.945 | 7.027 | 40.054 | 0.9756 |
| SRDNet(Ours) | | 0.996 | 2.756 | 42.902 | 0.9813 |

3) Results on Chikusei Dataset

The Chikusei dataset comprises 128 channels. To evaluate the performance of each algorithm under high upsampling factors, we selected upsampling factors of 4 and 8. The Fig. 12 displays the reconstructed images and spectral fit diagrams. The outcomes obtained with EDSR [47], GDRRN [9], and Bicubic methods are indistinct, whereas the SRDNet algorithm successfully reconstructs the primary structural features and spectral information of the image. Table IV presents the average performance of all compared algorithms on the four test images. It is evident that SRDNet outperforms all other algorithms, except for EUNet [28] when the upsampling factor is 4. However, as discussed earlier with the previous datasets, the EUNet [28] algorithm does not perform well when dealing with a large number of images. Furthermore, the computationally intensive operations in the network architecture of the MCNet [14] can significantly prolong the convergence time of the model. The limited number of images in the Chikusei dataset also impacts the performance of algorithms. To achieve models with better generalization, a larger number of samples are required for training. Although EDSR [47] networks demonstrate good performance in RGB image SR, its SAM index is relatively poor compared to single hyperspectral algorithms.

TABLE IV
DIFFERENT APPROACHES FOR THE 4 PQIS PAIRS OF 5 TEST IMAGES ON THE CHIKUSEI DATASET. RED AND BLUE SHOW THE BEST AND SECOND BEST PERFORMANCE RESPECTIVELY.

| Methods | Scale | CC↑ | SAM↓ | PSNR↑ | SSIM↑ |
|---|---|---|---|---|---|
| Bicubic | | 0.883 | 4.317 | 33.869 | 0.8135 |
| GDRRN[9] | | 0.975 | 3.324 | 36.767 | 0.9049 |
| 3D-FCNN[12] | | 0.978 | 3.201 | 37.363 | 0.9179 |
| EDSR[47] | ×4 | 0.977 | 3.462 | 37.252 | 0.9173 |
| MCNet[14] | | 0.979 | 3.009 | 37.657 | 0.9196 |
| EUNet[28] | | 0.981 | 2.762 | 38.050 | 0.9301 |
| SRDNet(Ours) | | 0.981 | 2.659 | 37.872 | 0.9286 |
| Bicubic | | 0.832 | 6.427 | 31.174 | 0.7438 |
| GDRRN[9] | | 0.945 | 4.922 | 33.381 | 0.8274 |
| 3D-FCNN[12] | | 0.945 | 4.894 | 33.502 | 0.8313 |
| EDSR[47] | ×8 | 0.938 | 5.534 | 32.960 | 0.8193 |
| MCNet[14] | | 0.947 | 4.842 | 33.606 | 0.8366 |
| EUNet[28] | | 0.949 | 4.532 | 33.751 | 0.8422 |
| SRDNet(Ours) | | 0.950 | 4.489 | 33.817 | 0.8455 |

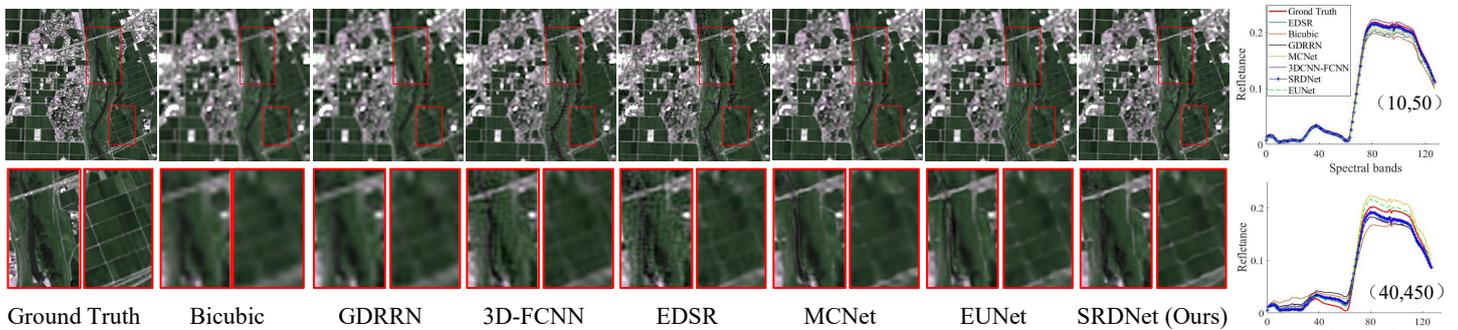

Ground Truth  Bicubic  GDRRN  3D-FCNN  EDSR  MCNet  EUNet  SRDNet (Ours)

**Fig. 12.** Reconstructed the *Chikusei_test_2* image of in Chikusei dataset with spectral bands 55-45-55 as R-G-B when the upsampling factor is 8. The last column is spectral fits at pixel points (10, 50), and (40, 450), respectively.

*C. Ablation Study*

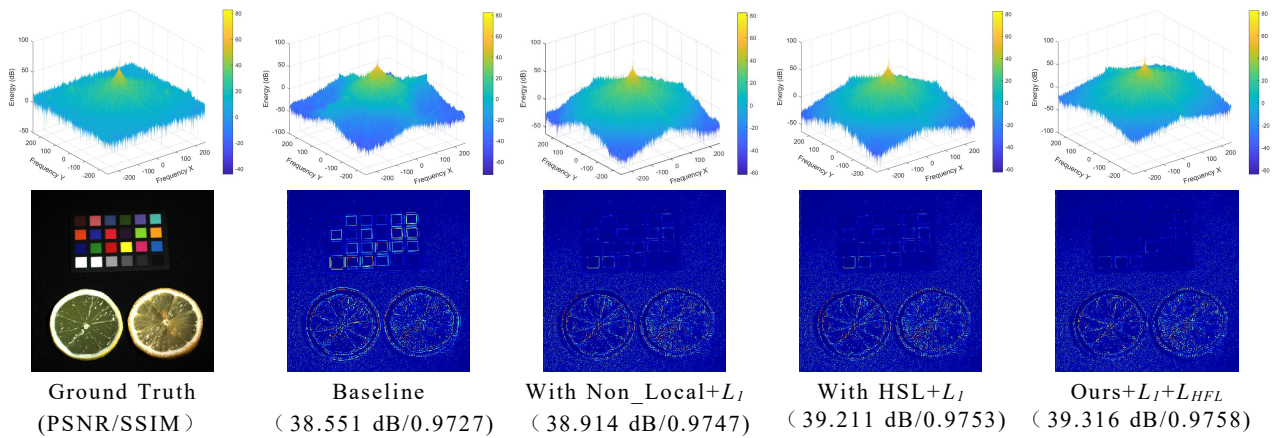

Ground Truth (PSNR/SSIM)  Baseline (38.551 dB/0.9727)  With Non_Local+$L_1$ (38.914 dB/0.9747)  With HSL+$L_1$ (39.211 dB/0.9753)  Ours+$L_1$+$L_{HFL}$ (39.316 dB/0.9758)

**Fig. 13.** The first row illustrates the spectrum power comparison of the *fake_and_real_lemon_slices_ms* in the CAVE when an upsampling factor is 3. In the graph, the data from the fifth band has been selected for processing. The spectral power of this band's data has then been calculated using FFT (Fast Fourier Transform) and is displayed in units of dB (decibels). The second row displays the absolute error.

As explained in subsection 3, our method has four main components, where the PAM block is the main part of the feature extraction network. We call the network without the PAM module the baseline. In this subsection, we study the effect of different combinations of PAM on model performance. Table V shows the ablation studies for these combinations at the upsampling factor of 3 on the CAVE dataset (fake_and_real_lemon_slices_ms). Specifically, when we remove the PAM module from the network, the performance drops for the same number of parameters. When the PAM module has only 3D units, the results are not very satisfactory. This is because the network pays too much attention to spectral information and loses spatial resolution. When the PAM has both 2D and 3D units, the performance is much better than only 2D units. Additionally, in our method, a certain degree of overfitting was observed when designing four 3D units. This could potentially be due to the limited number of images.

TABLE V
ABLATION STUDIES ON COMPONENTS, WHERE 'W/O' INDICATES THAT THIS MODULE IS NOT INCLUDED.

| Module | Component | CC↑ | SAM↓ | PSNR↑ | SSIM↑ |
|---|---|---|---|---|---|
| 3D uint(×) 2D uint(×) | Baseline | 0.983 | 2.884 | 42.469 | 0.9847 |
| 3D uint(√) 2D uint(√) | Two 3D uint | 0.987 | 2.741 | 42.925 | 0.9863 |
|  | Three 3D uint | 0.988 | 2.672 | 42.944 | 0.9869 |
|  | Four 3D uint | 0.987 | 2.679 | 42.937 | 0.9867 |
| 3D uint(√) 2D uint(×) | Three 3D uint | 0.985 | 2.793 | 42.611 | 0.9848 |
| 3D uint(×) 2D uint(√) | IGM(√) HSL(×) | 0.986 | 2.776 | 42.739 | 0.9859 |
|  | IGM(×) HSL(√) | 0.984 | 2.883 | 42.574 | 0.9850 |
|  | IGM(√) HSL(√) | 0.986 | 2.765 | 42.742 | 0.9861 |
| w/o HFL | $L_1$ | 0.988 | 2.681 | 42.913 | 0.9869 |
|  | $L_1+L_{HFL}$ | 0.988 | 2.672 | 42.944 | 0.9869 |

Fig. 13 shows that the absolute error of using the $L_{HFL}$ loss is lower than that of using only the $L_1$ loss. And even compared to the classic non-local attention mechanism (Non_Local), our proposed HSL has some advantages, and the absolute error information is lower than the latter. To better assess the HSL module we designed, we employ Fast Fourier Transform (FFT) to calculate power values at each frequency point as demonstrated in Eq. (39).

$$P\_x = 10 * \log^{10}(|f\_x|^2) \quad (39)$$

Where the P_x denotes power values at each frequency. f_x is the result of Fourier Transform shifted to zero frequency. The 10*log10 operation represents taking dB value of the energy spectrum, accentuating the differences between high and low frequencies.

Finally, when we combine all components into the model, we can see that it performs better than any other combination in all three aspects (see the Fig. 13). Thus, these analyses show that each component of SRDNet helps the model learn and optimize.

## V. CONCLUSION

We propose a hyperspectral image super-resolution algorithm via a dual-domain network (SRDNet) in this paper. Our method uses a hybrid convolution of 2D and 3D units with progressive upsampling to capture more spatial information while learning spectral information. Unlike previous work, a dual-domain learning network is designed. We use the spectral self-attention mechanism to select important spectral information adaptively, and the network can obtain fine and smooth pixel-level features. Moreover, to address the visual discrepancy caused by the pixel-level loss, a frequency loss is adopted to narrow the frequency domain difference between the reconstructed image and the ground truth. The effectiveness of various modules in enhancing spatial information and spectral coherence has been verified in ablation studies.

Visual analysis and quantitative experiments on three common hyperspectral datasets demonstrate that the SRDNet achieves excellent performance in HSI reconstruction at both pixel and frequency levels. Our algorithm achieves optimal results on several common objective metrics, and the images are perceptually closer to the ground truth than other methods.